\pdfoutput=1

\documentclass[11pt]{article}

\usepackage[preprint]{acl}

\usepackage{times}
\usepackage{latexsym}

\usepackage[T1]{fontenc}

\usepackage[utf8]{inputenc}

\usepackage{microtype}

\usepackage{inconsolata}

\usepackage{graphicx}
\usepackage{subcaption}

\usepackage{colortbl}

\usepackage{booktabs}
\usepackage{multirow}
\usepackage{pgfplots}
\usepackage{pgfplotstable}
\usepackage{filecontents}

\usepackage{algorithm}
\usepackage{algpseudocode}
\usepackage{amsmath}

\usepackage{tcolorbox}

\usepackage{listings}

\lstdefinelanguage{json}{
    basicstyle=\ttfamily\footnotesize,
    numbers=left,
    numberstyle=\scriptsize,
    stepnumber=1,
    breaklines=true,
    frame=single,
    backgroundcolor=\color{gray!10},
    literate=
     *{0}{{{\color{blue}0}}}{1}
      {1}{{{\color{blue}1}}}{1}
      {2}{{{\color{blue}2}}}{1}
      {3}{{{\color{blue}3}}}{1}
      {4}{{{\color{blue}4}}}{1}
      {5}{{{\color{blue}5}}}{1}
      {6}{{{\color{blue}6}}}{1}
      {7}{{{\color{blue}7}}}{1}
      {8}{{{\color{blue}8}}}{1}
      {9}{{{\color{blue}9}}}{1}
      {:}{{{\color{red}:}}}{1}
      {,}{{{\color{red},}}}{1}
      {"}{{{\color{orange}"}}}{1},
}

\title{Explainable Compliance Detection with Multi-Hop Natural Language Inference on Assurance Case Structure}

\author{Fariz Ikhwantri \\
  Simula Research Laboratory \\
  Oslo, Norway \\
  \texttt{fariz@simula.no} \\\And
  Dusica Marijan \\
  Simula Research Laboratory \\
  Oslo, Norway \\
  \texttt{dusica@simula.no} \\}

\pgfplotsset{compat=1.18} 

\begin{document}
\maketitle
\begin{abstract}



Ensuring complex systems meet regulations typically requires checking the validity of assurance cases through a claim-argument-evidence framework. Some challenges in this process include the complicated nature of legal and technical texts, the need for the model explanations, and limited access to assurance case data.
We propose a compliance detection approach based on Natural Language Inference (NLI): \textbf{EX}plainable \textbf{C}omp\textbf{L}iance detection with \textbf{A}rgumentative \textbf{I}nference of \textbf{M}ulti-hop reasoning (\textbf{EXCLAIM}). We formulate the claim-argument-evidence structure of an assurance case as a multi-hop inference for explainable and traceable compliance detection. We address the limited number of assurance cases by generating them using LLMs. We introduce metrics that measure the generated assurance cases' coverage and structural consistency. We demonstrate the effectiveness of the generated assurance case from GDPR requirements in a multi-hop inference task as a case study. 
Our results highlight the potential of NLI-based approaches in automating the regulatory compliance process.
\end{abstract}


\section{Introduction}




Compliance detection is a task to ensure that complex systems such as software adhere to established standards, regulations, and best practices~\citep{saeidi-etal-2021-cross, castellanos2022compliance}. Compliance detection aims to identify potential violations early in the development lifecycle, reducing legal risks and costly rework~\citep{duvall2007continuous}. 


\begin{figure}[ht]
    \centering
    \includegraphics[width=1.0\linewidth]{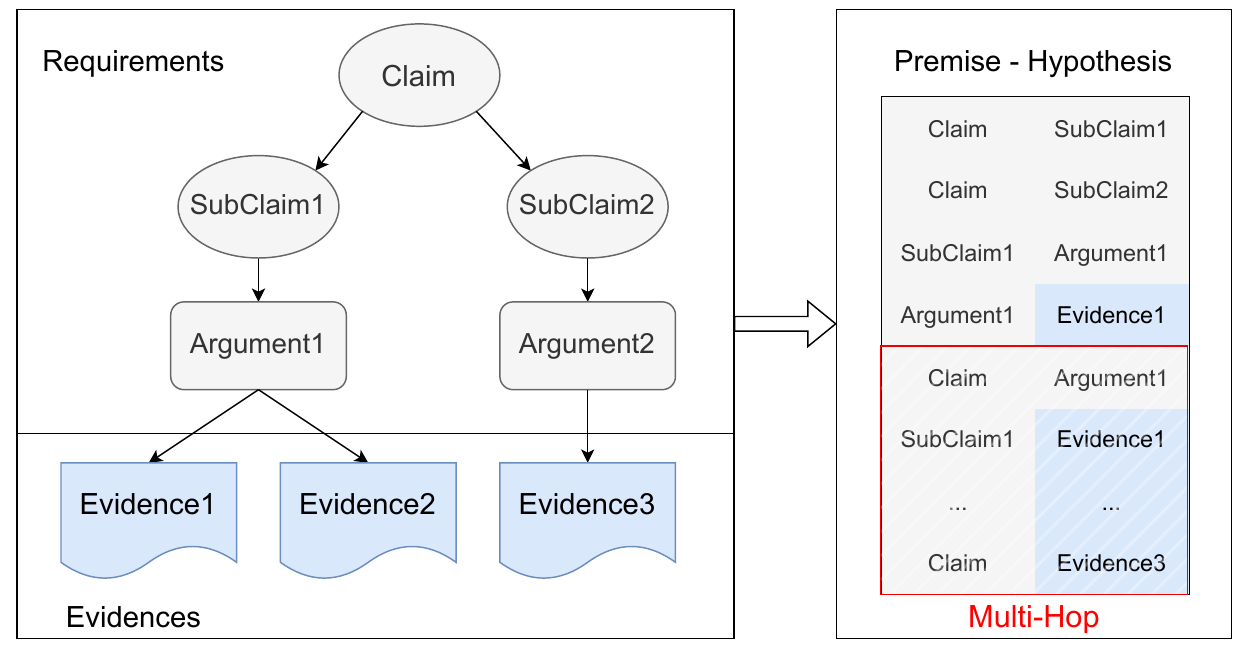}
    \caption{Claim-Argument-Evidence structure of an assurance case can be represented as NLI with direct and multi-hop (indirect) connections}
    \label{fig:multi-hop-nli}
\end{figure}

However, ensuring compliance is challenging due to the complexity of legal documents. Navigating privacy policy documents requires two sets of unique expertise in legal and technical domains. This process can be tedious for a legal expert, an auditor, or an engineer. This motivates many experts to automate the process of compliance detection. However, automating this process also means the AI system needs to be transparent. This demand makes it challenging to adapt to the black-box AI system. 


Previous study~\citep{10167495} uses inherently interpretable rule-based graph matching, but it still struggles with linguistic variability, ambiguous legal phrasing, and scalability to new regulations. A recent study~\citep{azeem2024multi} explores statistical machine learning and language model methods for the compliance detection task as text classification. However, it does not consider transparency and explainability factors such as model interpretability and reasoning.

To address the transparency and robustness factors of detecting risk and safety, experts use assurance cases~\citep{rhodes2010software, MANSOUROV201123} as a structured argument approach to audit complex software products in aircraft or autonomous vehicles to standard requirements. This is to ensure compliance with legal and technical standards. An assurance case is a systematic collection of arguments and supporting evidence to demonstrate a certain level of confidence that a product meets particular claims related to its requirements~\citep{Netkachova2015ToolSF, PIOVESAN2017113}. These assurance cases can be created and updated throughout the development lifecycle~\citep{Ross2022EngineeringTS}.


\begin{figure*}[ht]
    \centering
    \includegraphics[width=1.0\linewidth]{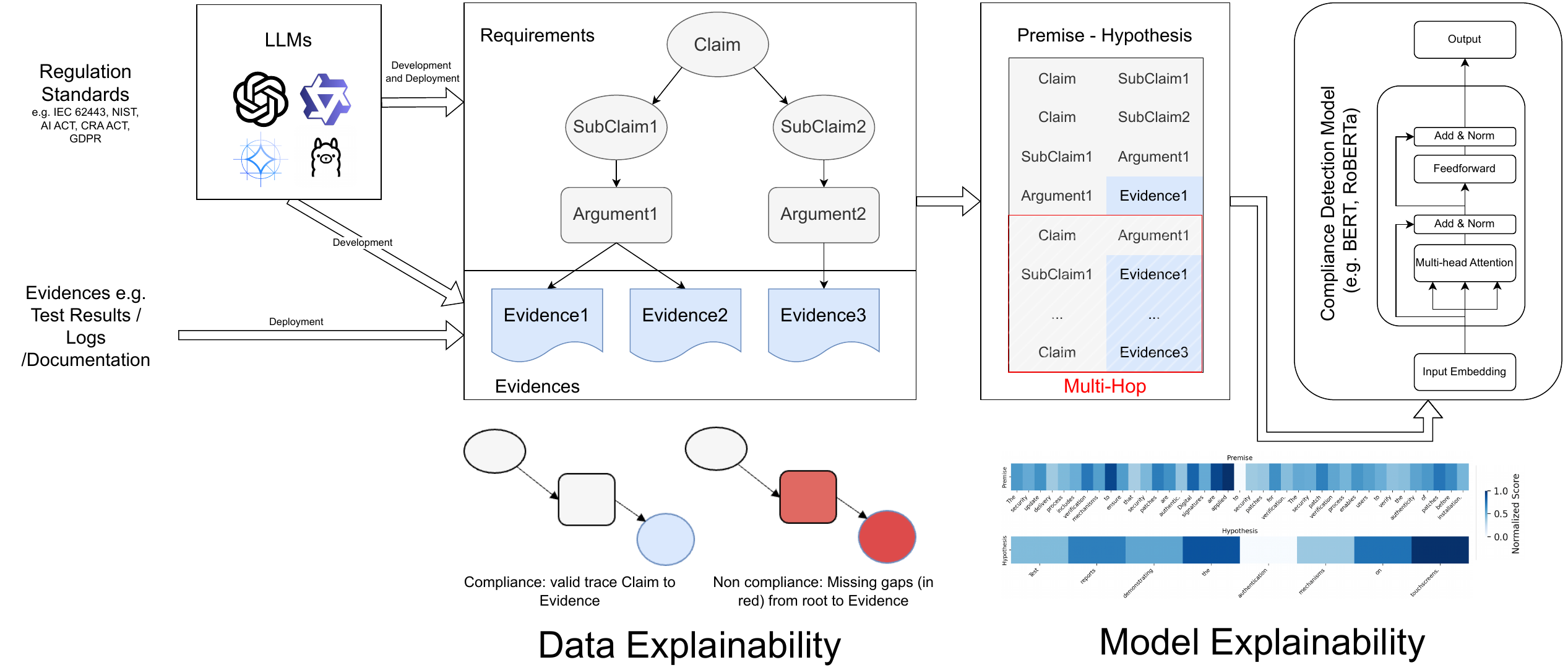}
    \caption{The framework of our explainable compliance detection with multi-hop NLI on Claim-Argument-Evidence (CAE) structure of an Assurance Case.}
    \label{fig:framework}
\end{figure*}



Despite its significance, obtaining and using assurance case data presents important challenges. One of the foremost challenges is the lack of publicly available datasets. 
Another challenge is the high cost of acquiring and annotating assurance case data. Developing and validating assurance cases requires significant domain expertise, making large-scale annotation expensive and time-consuming.


Recently \citet{ODU2025112353} proposed to instantiate the Assurance Case patterns of Goal Structuring Notation (GSN)~\citep{kelly2004goal} with Large Language Models (LLMs). They assess the data produced by LLMs by utilising shallow lexical metrics from the text generation task. \citet{CHEN2025103288} used LLMs to create Assurance Cases and evaluated them using constraint solvers from Formal Methods~\citep{JAFFAR1994503}. However, constraint solvers cannot handle every assurance case completely, as they frequently involve qualitative reasoning, contextual relationships, and uncertainties that extend beyond strict logical or mathematical boundaries. 


One possible approach is to use deductive reasoning~\citep{rushby2015understanding} on the assurance case. Deductive reasoning in the assurance case derives conclusions from structured claims and arguments from evidence~\citep{bloomfield2009safety}. This reasoning can be framed in the Natural Language Inference (NLI) task~\citep{dagan2005pascal, bowman-etal-2015-large} to determine if a hypothesis follows from a premise. If there is a set of premises, multi-hop inference~\citep{jansen-2018-multi, yang-etal-2018-hotpotqa} is needed for reasoning over interconnected claims and evidence. This makes multi-hop inference suitable for assurance cases that link high-level system properties to low-level evidence, like test results.






 This paper proposes \textbf{EX}plainable \textbf{C}omp\textbf{L}iance detection with \textbf{A}rgumentative \textbf{I}nference of \textbf{M}ulti-hop reasoning (\textbf{EXCLAIM}) based on Assurance Case. 
 Figure~\ref{fig:framework} illustrates our workflow that explains this paper's contribution, which provides a summary of this paper. 
 From the figure, we utilise pre-trained transformer models~\citep {devlin-etal-2019-bert} to learn premise-hypothesis pairs for deductive multi-hop reasoning, enabling compliance detection. This reasoning process can detect non-compliance by providing a trace as an explanation of gaps between the main claim of fulfilled requirements and the evidence. 
 
Specifically, our contributions address challenges in (i) explainable compliance detection, (ii) advanced reasoning and (iii) the lack of assurance case data as follows:
\begin{itemize}
    \item \textbf{In terms of the model explainability}, we analyse the faithfulness of interpretation methods applied to the NLI models for the compliance detection task. We consider four interpretation methods: gradient, Integrated gradients, LIME, and SHAP.
    \item \textbf{In terms of the data explainability and reasoning}, we formulate the claim-argument-evidence structure from the assurance case as a discourse framework. This framework can be utilised to develop a multi-hop inference model for the explainable tracing of the compliance detection model's output (Figure~\ref{fig:multi-hop-nli}).
    \item \textbf{To address the lack of assurance case data}, we propose evaluation methods to measure the assurance case generated by large language models (LLMs) to validate the reliability of instance-level consistency and structural coherence. We conduct a study by generating Claim Argument Evidence Assurance Case data with open source and proprietary LLMs from GDPR requirements as the context prompt from a previous study~\citep{10167495}. We analyse the generated Assurance Case data by using our proposed metrics. 
\end{itemize}





\section{Related Work}

 \paragraph{Compliance detection in Requirements}
Compliance with legal regulations and industry standards has become necessary as software systems, such as finance and healthcare, become more integrated into people's daily lives. To ensure compliance, software requirements must be verified to align with applicable laws, regulations, standards, and organisational policies, for example, in data protection laws (e.g., GDPR, CCPA), cybersecurity (e.g., IEC 62443, NIST 800-53), and safety-critical standards (e.g., HIPAA). This process is known as compliance detection.

Recent studies have extensively evaluated statistical machine learning and Language model approaches as text classification tasks for compliance detection~\citep{azeem2024multi} on GDPR. Their study reveals that text classification methods outperform traditional rule-based graph matching~\citep{10167495}, offering greater flexibility and adaptability in handling various types of data. However, they do not address crucial aspects of explainability, i.e. the reasoning processes behind the model predictions. This lack of focus on explainability raises concerns about the transparency and trustworthiness of the employed automated system.
\paragraph{Explainable AI}



Explanation of model prediction provides the rationale behind a model's decision. The rationale can be obtained by analysing the contribution of individual features to the decision-making process~\citep{molnar2025}. 
Model explanations can be either inherent or post-hoc. Shallow models like linear regression are inherently interpretable, whereas deep learning and LLMs typically require post-hoc methods such as gradient-based~\citep{simonyan2014deep, sundararajan2017axiomatic} or input-perturbation approaches~\citep{ribeiro2016should, lundberg2017unified}. In addition, the explanation needs to be evaluated by plausibility~\citep{hollenstein-beinborn-2021-relative, eberle-etal-2022-transformer, IKHWANTRI2023103195, ikhwantri-etal-2024-analyzing} or faithfulness~\citep{atanasova-etal-2020-diagnostic, bastings-etal-2022-will}.

\section{Methodologies}

This section describes our proposed methodologies for forming compliance detection as a natural language inference task and extending it to multi-hop reasoning from the assurance case structure.

\subsection{Compliance Detection as Natural Language Inference}

\begin{table*}[ht]
\centering
\footnotesize
\begin{tabular}{lp{5cm}p{5cm}ll}
\toprule
Req. & Premise (Req) & Hypothesis (DPA) & DPA-ID & Label \\
\midrule
 R18 & The processor shall assist the controller in consulting the supervisory authorities prior to processing ... to mitigate the risk. (Art. 28(3)(f), Art. 36) & At controller’s request and at the controller’s reasonable expense on a time and materials basis, ... fulfilling any ... Applicable Data Protection Law. & Online 100 &  entailment \\
\bottomrule
\end{tabular}
\caption{Compliance Detection as NLI task where GDPR requirement is the premise and Data Processing Agreement is the hypothesis}
\label{fig:example-real}
\end{table*}

Natural Language Inference (NLI) is a core task in natural language understanding that involves determining the logical relationship between two textual statements~\citep{dagan2005pascal}. We propose compliance detection by devising the requirements as premises and evidence as hypotheses. This method allows the model to learn from the verbalisation of requirements text instead of discrete classes compared to previous studies, as text classification~\citep{azeem2024multi}. 

For example, in Table~\ref{fig:example-real}, the inputs are GDPR requirements, which served as the premise, and the Data Processing Agreement (DPA) served as the hypothesis. The entailment label means requirements and DPA belong to the same GDPR requirement class and comply, while the non-entailment label means requirements and DPA do not belong to the same GDPR requirement class, i.e. neutral or contradict. 

This is similar to requirements mapping~\citep{Fazelnia2024LessonsFT} for requirements classification approaches, which are based on few-shot~\citep{wang2021entailment} or zero-shot scenario~\citep{sainz-etal-2021-label}. The output is similar to binary classification approaches~\citep{azeem2024multi}. However, NLI only requires a model instead of $n$ models for binary classification, where $n$ is the number of requirements. 


\subsection{Assurance Case Structure as Multi-hop Inference} 

In the assurance case, the claim-argument-evidence (CAE) structure could be leveraged to construct premise-hypothesis text pairs in the NLI task. These CAE trees provide a formulation for deductive inference, allowing for sequential reasoning, where multiple premises collectively support a conclusion. 

We construct multi-hop premise-hypothesis pairs where a hypothesis is inferred from the indirect connection of a sub-claim to evidence or the main claim to an argument sub-claim. An example can be seen in Figure \ref{fig:multi-hop-nli}. From the figure, the parent node in the assurance case structure serves as the premise. The child node, which builds upon or supports the parent, becomes the hypothesis. These structured relationships are extracted from assurance cases and transformed into a format suitable for training NLI models. This approach enables our systems to perform more complex inferences in a simple NLI model as a strong baseline.

\section{Dataset}

\subsection{GDPR-DPA}

We use the compliance detection dataset from a previous study~\citep{azeem2024multi} for the NLI task on a real-world human-annotated dataset\footnote{\href{https://zenodo.org/records/11047441}{https://zenodo.org/records/11047441}}. We called this dataset the GDPR-DPA from the General Data Protection Regulation-Data Processing Agreement (GDPR-DPA). The details of the dataset statistics are presented in Table~\ref{tab:combined_dataset_stats}. 
The number of instances in the training set is larger than the original data due to pairing of negative samples between non-matching premises (requirements) and the hypothesis (DPA). 
We set the sampling rate to 0.1 to add negative pairs within the iteration of unique hypotheses times the number of unique premises.

\begin{table}[ht]
    \centering
    \footnotesize
    \begin{tabular}{lrr}
        \toprule
        \textbf{Statistics} & \textbf{Train} & \textbf{Test} \\
        \midrule
        Instances          & 362317 & 1511  \\
        Unique ID           & 61     & 8    \\
        Unique premise      & 45     & 45   \\
        Unique hypothesis   & 9820   & 1364  \\
        Unique target       & 45     & 17   \\
        \bottomrule
    \end{tabular}
    \caption{Dataset Statistics for NLI-Based Compliance Detection on GDPR-DPA Dataset: Train and Test}
    \label{tab:combined_dataset_stats}
\end{table}

\subsection{LLM Generated Data}

We use the GDPR requirements compiled by~\citet{10167495} for the prompt to generate assurance cases. There are 45 requirements, but only 20 are considered for compliance detection~\citep{azeem2024multi}. We generated an Assurance case using three proprietary LLMs, namely gpt3.5-Turbo-4k, gpt4o-8k and gpt-4o. From open source LLMs, we use models such as Llama3 (3.2-1B, 3.1-8B, 3.1-70B)~\citep{grattafiori2024llama}, Qwen2.5 (7B, 14B, 72B)~\citep{qwen2}, Gemma (7B, 9B, 27B)~\citep{team2024gemma}, and phi4~\citep{abdin2024phi}.


\subsection{Evaluating LLMs generated Assurance Case}

A recent study \citet{ODU2025112353} used Goal-Structured Notation to generate assurance cases from five products. This structure differs from the Claim-Argument-Evidence structure. In addition, their methods rely on the availability of gold assurance, which is not always available and is hard to obtain. Their work evaluates the LLMs' generated data using shallow lexical metrics such as the BLEU score from the machine translation task. This lexical metric does not give insight into the higher-level view of the generated data. 

\paragraph{Flat metric}

We proposed Flat metrics, which assess the assurance case at a high level without analysing textual content or internal relationships between elements. This approach provides a quantitative assessment based on the structural components of the generated assurance case. 



We measure the quantitative assessment by calculating the absolute difference for each element type, such as the main claim, sub-claim, argument claim, and evidence from two assurance cases of the same requirements. Flat metrics provide an initial assessment of coverage, balance of elements, and consistency. Concretely, we can detect possible redundancy across assurance cases because they focus on quantitative elements without regard for the structures.

\paragraph{Structured metric}

Assurance cases are structured arguments, often represented as graphs or trees. We need to consider the graph's structure to assess the quality of assurance cases generated by LLMs. This means analysing the hierarchical or graph-based relationships between claims, arguments, and evidence in an assurance case. 

One possible approach to measuring the similarity between two graphs is Graph Edit Distance (GED). GED is a measure of dissimilarity between two graphs. This metric quantifies the minimum number of edit operations required to transform one graph into another. 
A lower value indicates higher similarity, while a higher value indicates lower similarity. The value of zero indicates the two graphs are isomorphic.

\paragraph{Data Analysis}

We divide the Flat and Structured metric analysis into \textbf{intra-model} and \textbf{inter-model} comparison to evaluate the claim-argument-evidence (CAE) structure generated by LLMs. 

\textbf{Intra-model} comparison evaluates how consistently a single model performs across multiple inferences, similar to intra-annotator agreement in human data labelling. On the other hand, \textbf{inter-model} comparison examines the agreement between the outputs of different models.

Table~\ref{tab:flat-metric} shows the difference between four types of instances generated by LLMs. In flat metric, generated instances of type MainClaim, SubClaim and ArgumentClaim at the instance levels have lower differences, which means high consistency. Overall, models tend to diverge in evidence instances. This means that efforts can be made to check the validity of the assurance case structure by checking the coverage of evidence rather than the argument and SubClaim.

In structured metrics, Gemma2 models(v2-9b-it, v2-27b-it) and phi4 generate a consistent structure, which is shown by their intra-model structured metrics (GED) results. The lowest inter-model value is a 3.26 GED score between the phi4 and chatGPT4o models. 



\begin{table}[ht]
  \centering
  \footnotesize
  \begin{tabular}{lcc}
    \toprule
    Type & Intra-model & Inter-model \\
    \midrule
    MainClaim        & 0.00   & 0.00    \\
    SubClaim         & 0.08   & 0.24   \\
    ArgumentClaim    & 0.16   & 0.32   \\
    ArgumentSubClaim & 0.37   & 0.90   \\
    Evidence         & 0.89   & 2.90   \\
    \bottomrule
  \end{tabular}
  \caption{Flat-metrics: Absolute Count Difference for Intra- and Inter-Model Agreement}
  \label{tab:flat-metric}
\end{table}

\begin{figure}[ht]
    \includegraphics[width=1.0\linewidth]{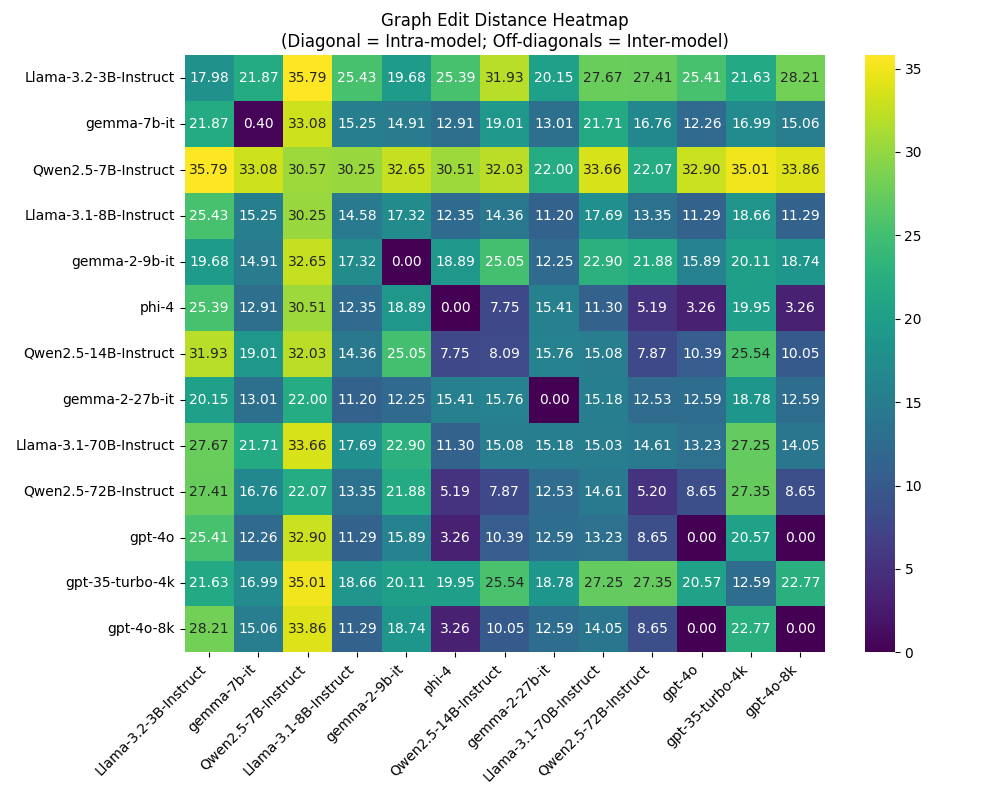}
    \caption{Graph Edit Distance between intra and inter-model generated assurance case structure.}
    \label{fig:enter-label}
\end{figure}

Table~\ref{tab:ac_stats} describe the dataset statistics details of final LLMs generated assurance cased as multi-hop inference task.

\begin{table}[ht]
\centering
\footnotesize
\begin{tabular}{lrr}
\toprule
 & Train & Test \\
 \midrule
All & 103838 & 12875 \\
\#1\_hop & 12241 & 3150 \\
\#2\_hop & 10725 & 2723 \\
\#3\_hop & 9136 & 2286 \\
\#4\_hop & 6192 & 1495 \\
\bottomrule
\end{tabular}
\caption{Dataset statistics of Generated Assurance Case by LLMs in Evaluating LLM Generated Data}
\label{tab:ac_stats}
\end{table}

\section{Experiments Setup}

In this section, we explain the experiment setup for evaluating compliance detection as a natural language inference task on the GDPR-DPA dataset and the multi-hop inference from the assurance case data generated by LLMs.




 


\subsection{Research Questions} In this paper, we aim to answer the following questions:
\begin{description}
    \item[RQ1] How does compliance detection as NLI performance compare to the previous approach?
    \item[RQ2] How does the model's prediction on compliance detection output affect its explanation? 
    \item[RQ3] What is the model's performance in compliance detection on the unseen requirements?
    \item[RQ4] Does the synthetic reasoning improve the model's faithfulness in detecting compliance with the unseen requirements?
\end{description}

\subsection{Evaluating Explanation – Faithfulness Metrics}

Faithfulness refers to the degree to which the model relies on the output of local explanation (rationales) to make its prediction. It ensure the provided rationales are not just plausible but also truthful representations of the model's inner workings~\citep{jacovi-goldberg-2020-towards}. There are two faithfulness metrics we consider in this work.


\paragraph{Comprehensiveness} evaluates whether all necessary rationales for making a prediction have been chosen~\citep{deyoung-etal-2020-eraser, atanasova-etal-2020-diagnostic}. 
This can be done by creating a contrasting example for x, denoted as $\widetilde{x}$, which is x without the predicted rationales r. In a classification context, let $m{(x)}_j$ represent the original model prediction for the output class $j$. Next, we examine the predicted probability from the model for the same class after eliminating the supporting rationales. The model should demonstrate reduced confidence in its prediction once the rationales are removed from x. This can be expressed as $\mathrm{Compr}=m\left(x\right)_j-m\left(x|r\right)_j$, where ${(x|r)}_j$ is the sentence x where tokens in r are removed. 
This metric indicates that a greater difference between the original sentence and the modified version, excluding the specified tokens, is preferable. A higher value signifies more significant alterations and, consequently, a more effective reduction of the input while retaining essential meaning. 



\paragraph{Sufficiency} evaluates whether the identified rationales are adequate for the model to make accurate predictions~\citep{deyoung-etal-2020-eraser, atanasova-etal-2020-diagnostic}. It measures the model's confidence in its prediction when only the rationales are retained, compared to the original input. The sufficiency metric is defined as $\mathrm{Suff}=m\left(x\right)_j-m\left(r\right)_j$, where $(m\left(x\right)_j)$ represents the model's original prediction confidence for the output class $(\ j\ )$, and  $(m\left(r\right)_j)$  represents the model's prediction confidence when only the rationales $(\ r\ )$ are retained. In this metric, a value close to zero is better, indicating that the model can still make confident predictions with just the rationales, demonstrating their Sufficiency. 




\section{Experiment Scenarios}

\subsection{NLI-Based Compliance Detection}

To address the challenges of legal text ambiguity, domain specificity, and evolving regulations in compliance detection, we explore transfer learning methods and compare the results with a previous study~\citep{azeem2024multi} to answer \textbf{[RQ1]}.
In this scenario, we compare three different transfer learning algorithms, namely Fine-tuning (FT), Linear-Probing-Fine-tuning (LP-FT)~\citep{kumarfine} and Low-Rank Adaptation (LoRA)~\citep{hu2022lora}. 


Our approach considers two main pre-trained language models (LM) categories: general-purpose and domain-specific legal models. For general domain, we consider BERT-large~\citep{devlin-etal-2019-bert} (BERT), RoBERTa-large~\citep{liu2019roberta} (RoBERTa). For domain-specific legal models, we consider Legal-BERT-base~\citep{chalkidis-etal-2020-legal} (Legal-BERT) as the counterpart of BERT and Legal-RoBERTa~\citep{chalkidis-etal-2023-lexfiles} as the counterpart of RoBERTa. In addition, we also explore the decoder-based LM, such as GPT-2-XL\footnote{\href{openai-community/gpt2-xl}{openai-community/gpt2-xl}} and Llama-3.2-1B~\citep{grattafiori2024llama}, to classify the entailment as a text generation baseline. 




To answer \textbf{[RQ2]}, we explore how model faithfulness influences compliance detection output. 
We evaluate four post-hoc explanation methods, such as Gradient (Grad)~\citep{simonyan2014deep}, Integrated gradient (IG)\citep{sundararajan2017axiomatic}, LIME~\citep{ribeiro2016should} and Partition SHAP (Part SHAP)~\citep{lundberg2017unified}. In this paper, we use the ferret library~\citep{attanasio-etal-2023-ferret} implementation to measure the model's faithfulness. The faithfulness metric output is a score of the area over the perturbation curve (AOPC). AOPC is obtained by aggregating the mean of the Comprehensiveness or sufficiency score over bins of 10\% which are applied to the models.

\subsection{Multi-Hop Inference}

We evaluate whether transformer-based language models can perform deductive inference on CAE tree-based assurance cases. We explicitly append indirect premises to the model input as a simple baseline. Appending indirect premises allows the models to perform multi-hop inference. We aim to assess whether explicit intermediate reasoning steps improves inference performance compared to a model that only sees the indirect premise and hypothesis. 


In this scenario, we split the train and test of the generated assurance case based on the requirement categories to answer \textbf{[RQ3]}. From the 19 requirements, we used 15 unique requirements for training data and requirements for test data. We use the same models but decided to focus on only fine-tuning to focus on analysing the explanation improvement to answer \textbf{[RQ4]}.




\section{Results and Discussion}

\begin{table}[htbp]
\centering
\scriptsize
\begin{tabular}{llcccc}
\toprule
Model & Method & pre & rec & f1 & f2 \\
 \midrule
\multicolumn{4}{l}{\cite{azeem2024multi}}   \\
\midrule
BERT* & Binary-FT & .82 & .82 & .82 & .82 \\
    & Multi-FT & .83 & .85 & .84 &	.84 \\
\midrule
RoBERTa* & Binary-FT & .84 & .63 & .74 & .66 \\
 & Multi-FT & .81 & .90 &  .85 &	.87 \\
\midrule
This Work & &  &  &  \\
\midrule
 & FT & .81 & .85 & .83 & .83 \\ 
BERT & LP-FT & .91 & .76 & .81 & .79 \\ 
 & LoRA & .81 & .82 & .81 &.82 \\ 
 \midrule
 & FT & .90 & .83 & .86 &.84 \\ 
Legal-BERT & LP-FT & .89 & .75 & .80 & .77 \\ 
 & LoRA & .75 & .56 & .58 & .59 \\ 
 \midrule
 & FT & .89 & .72 & .78 & .75 \\ 
RoBERTA & LP-FT & .50 & .50 &   .50 & .50 \\ 
 & LoRA & .80 & .86 & .82 & .85 \\ 
 \midrule
 & FT & .50 & .50 & .50 & .50 \\
Legal-RoBERTA & LP-FT & .50 & .50 & .50 & .50 \\
 & LoRA & .91 & .86 & .88 &.87 \\ 
\midrule
GPT-2-XL (1.5B) & Zero-Shot & .49 & .48 & .48 & .48 \\
        & One-Shot & .53 & .58 & .55 & .57 \\
\midrule
Llama-3.2 (1B) & Zero-Shot & .49 & .49 & .49 & .49 \\
        & One-Shot & .52 & .53 & .52 & .53 \\
 \bottomrule
\end{tabular}
\caption{Pre-trained language model trained on Compliance detection as NLI task. Previous work use base model size (108M) vs ours mostly use large (334M) except for Legal-BERT}
\label{tab:finetune-exp}
\end{table}


\paragraph{NLI-Based Compliance Detection}

Table~\ref{tab:finetune-exp} shows compliance detection as NLI performance compared to previous approaches. We report the precision (\textit{pre}), recall (\textit{rec}), and \textit{f2}-score. We follow previous approaches for a fair comparison to report the f2-score because it puts more weight on recall than precision. The BERT-FT approach performs better than the BERT with binary classification fine-tuning (Binary FT). BERT with multi-class fine-tuning (Multi-FT) performs better than our best results. However, we do not find the decoder-based LM (GPT-2-XL and Llama-3.2-1B) performance satisfactory.

The recall performance between our BERT-FT and the previous approach, BERT with multi-class classification fine-tuning (Multi-FT), is competitive. Our approach is better than BERT binary classification fine-tuning (Binary-FT). The answer to \textbf{RQ1} "\textit{How does compliance detection as NLI performance compare to the previous approach?}" is that our best performing approach performs comparably to previous approaches (Binary-FT). However, our approach can provide better explainability and traceability to verbalisation requirements.


Table~\ref{tab:cor_inc_exp} shows comprehensiveness and sufficiency scores for each model and explainer combination. Overall, LIME methods perform best in terms of the comprehensiveness metric. Regarding the sufficiency metric, Legal-BERT-FT with Part-SHAP shows closer values to zero.

\begin{table*}[ht]
\centering
\footnotesize
\begin{tabular}{llcccccc}
\toprule
 &  & \multicolumn{3}{c}{AOPC Comprehensiveness} & \multicolumn{3}{c}{AOPC Sufficiency} \\
 \midrule
Model & Explainer & Overall & Correct & Incorrect & Overall & Correct & Incorrect \\
\midrule
\multirow{4}{*}{BERT-FT} & IG & .276(.26) & .606(.17)* & .168(.18) & -.009(.19) & .188(.19)* & -.073(.13) \\
 & Grad (x Input) & .215(.26) & .467(.26)* & .133(.20) & .047(.23) & .298(.24)* & -.035(.15) \\
 & LIME & .414(.25) & .694(.13)* & .323(.21) & -.058(.15) & .048(.15)* & -.093(.13) \\
 & PartSHAP & .367(.25) & .671(.13)* & .267(.20) & -.059(.14) & \textbf{.038(.14)}* & -.091(.13) \\
 \midrule
\multirow{4}{*}{Legal-BERT-FT} & IG & .137(.30) & .760(.14)* & .021(.13) & .032(.24) & .518(.17)* & -.059(.10) \\
 & Grad (x Input) & .072(.27) & .658(.16)* & -.038(.09) & .069(.31) & .752(.14)* & -.058(.10) \\
 & LIME & .210(.30) & .764(.15)* & .107(.19) & .018(.22) & .445(.17)* & -.062(.10) \\
 & PartSHAP & .243(.30) & .759(.15)* & .147(.21) & \textbf{.006(.19)} & .367(.19)* & -.061(.10) \\
 \midrule
\multirow{4}{*}{RoBERTa-LoRA} & IG & .450(.36) & .666(.15)* & -.044(.15) & .376(.38) & .599(.18)* & -.135(.17) \\
 & Grad (x Input) & .314(.34) & .509(.19)* & -.133(.15) & .474(.42) & .729(.16)* & -.112(.16) \\
 & LIME & \textbf{.629(.31)} & \textbf{.807(.12)*} & .218(.17) & .214(.30) & .374(.19)* & -.154(.16) \\
 & PartSHAP & .567(.36) & .784(.13)* & .066(.16) & .237(.31) & .412(.17)* & -.166(.16) \\
 \midrule
\multirow{4}{*}{Legal-RoBERTa-LoRA} & IG & .446(.30) & .587(.17)* & -.012(.14) & .392(.35) & .562(.17)* & -.167(.16) \\
 & Grad (x Input) & .298(.31) & .423(.23)* & -.107(.14) & .515(.41) & .720(.18)* & -.155(.15) \\
 & LIME & .672(.26) & .791(.13)* & .283(.21) & .154(.25) & .258(.17)* & -.187(.15) \\
 & PartSHAP & .612(.27) & .742(.12)* & .190(.20) & .177(.25) & .287(.15)* & -.183(.15) \\
 \bottomrule
\end{tabular}
\caption{Correct vs Incorrect model-explainer faithfulness metrics score. * denote significant difference with $p<.001$ from unpair random permutation testing.}
\label{tab:cor_inc_exp}
\end{table*}

Further analysis shows that correct model predictions generally have higher comprehensiveness scores than incorrect ones, suggesting that faithfulness is associated with predictive accuracy. While the sufficiency metric's on the correct scores should be closer to zero than the incorrect ones, we argue that it might not be a reliable indicator of completeness in legal compliance checking. Relying only on salient words can lead to ambiguous and misaligned hypotheses with real-world legal reasoning. However, the sufficiency metric might be more relevant to higher-level representation, such as when sentences or paragraphs are kept within documents instead of tokens.

Interestingly, we do not find any significant effect of in-domain pre-training on model faithfulness, indicating that domain-specific training may improve accuracy but does not necessarily enhance interpretability or explanation quality. 

The answer to our~\textbf{RQ2} "\textit{How does the model's prediction on compliance detection output affect its explanation?}" is that the model's correct prediction affects explanation on Comprehensiveness. In contrast, the model's incorrect prediction relates more to the sufficiency metric than the correct prediction.


\paragraph{Multi-Hop Inference}



\begin{figure}[htbp]
    \centering
    \footnotesize
    \resizebox{0.9\columnwidth}{!}{
    \begin{tikzpicture}
    \begin{axis}[
    width=12cm,
    height=8cm,
    xlabel={Hop},
    ylabel={Macro F1 Score},
    xtick={1,2,3,4},
    xticklabels={1, 2, 3, 4},
    ymin=0.70, ymax=1.00,
    legend pos=south east,
    grid=both,
    legend style={
        at={(0.5,-0.2)}, 
        anchor=north, 
        legend columns=2, 
        /tikz/every even column/.append style={column sep=1cm}, 
        font=\footnotesize
    },
  ]
    \addplot[mark=square*, thick, color=blue] coordinates {(1,0.96091) (2,0.90002) (3,0.83782) (4,0.78653)};
    \addlegendentry{BERT-large, wo\_chain}
    
    \addplot[mark=square*, thick, color=blue, dashed] coordinates {(1,0.96840) (2,0.96895) (3,0.96444) (4,0.95819)};
    \addlegendentry{BERT-large, chain}
    
    \addplot[mark=triangle*, thick, color=red] coordinates {(1,0.96789) (2,0.90290) (3,0.85385) (4,0.78455)};
    \addlegendentry{RoBERTa-large, wo\_chain}
    
    \addplot[mark=triangle*, thick, color=red, dashed] coordinates {(1,0.96991) (2,0.96405) (3,0.95497) (4,0.94829)};
    \addlegendentry{RoBERTa-large, chain}
    
    \addplot[mark=o, thick, color=green!70!black] coordinates {(1,0.94754) (2,0.88127) (3,0.82889) (4,0.74475)};
    \addlegendentry{Legal-BERT, wo\_chain}
    
    \addplot[mark=o, thick, color=green!70!black, dashed] coordinates {(1,0.94508) (2,0.95410) (3,0.95210) (4,0.94122)};
    \addlegendentry{Legal-BERT, chain}
    
    \addplot[mark=diamond*, thick, color=orange] coordinates {(1,0.97193) (2,0.92647) (3,0.88732) (4,0.80400)};
    \addlegendentry{Legal-RoBERTa, wo\_chain}
    
    \addplot[mark=diamond*, thick, color=orange, dashed] coordinates {(1,0.95592) (2,0.96168) (3,0.95854) (4,0.94047)};
    \addlegendentry{Legal-RoBERTa, chain}
    
  \end{axis}

\end{tikzpicture}
}
\caption{F1 Scores for Different Models Across Various Hop Difficulties}
\label{fig:f1_scores}
\end{figure}
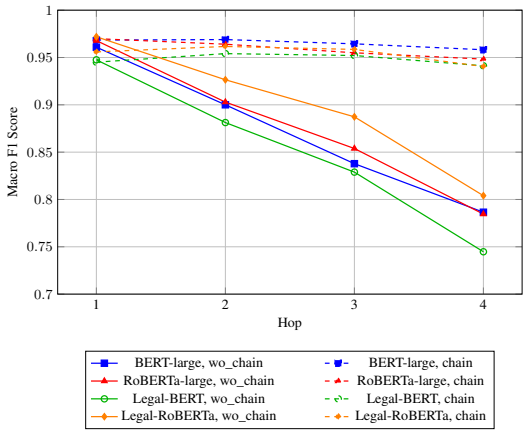

Figure~\ref{fig:f1_scores} shows results from our experiment in multi-hop inference from SACs. From the figure, we can observe the accuracy drop as the number of hops increases for the indirect model (wo\_chain). This result suggests that deep reasoning structures are challenging for NLI models without explicit intermediate steps. The answer to \textbf{RQ3} "\textit{What is the model's performance in compliance detection on the unseen requirements?}" is that the explicit model's performance is better on unseen requirements than the implicit model's. 


\begin{figure}[ht]
    \centering
    \includegraphics[width=0.40\textwidth]{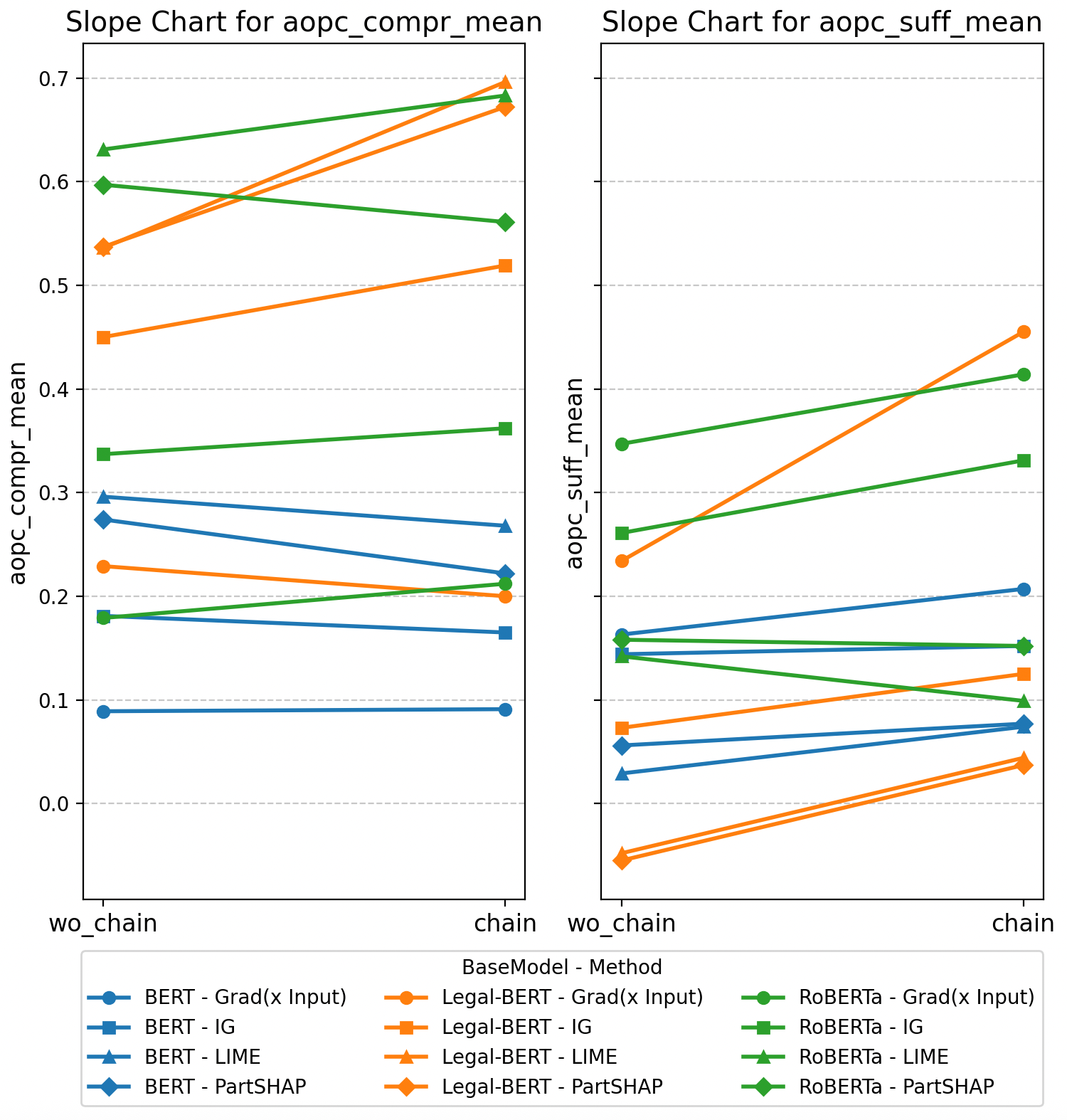}
    \caption{Slope Chart of AOPC Compr by Model and Method}
    \label{fig:slope_wo_chain_vs_chain}
\end{figure}

We analyse the post-hoc explanation of models with \textbf{(chain)} vs without chain \textbf{(wo\_chain)}. Figure~\ref{fig:slope_wo_chain_vs_chain} shows the faithfulness metrics difference between without chain and with chain. In the Comprehensiveness metric, ideally, \textbf{the score should increase from the model without the chain to the model with the chain}. These results were observed in the Legal-BERT model and RoBERTa. There are some exception however with the BERT model and RoBERTa models with PartSHAP. In the Sufficiency metric, ideally, \textbf{the score should decrease from the model without the chain to the model with the chain}. These results were only observed in the RoBERTa model with LIME and a slight decrease in the RoBERTa model with IG. The answer to \textbf{RQ4} "\textit{Does the synthetic reasoning improve the model's faithfulness in detecting compliance with the unseen requirements?}" is that the synthetic reasoning improves the model's Comprehensiveness (by increasing value) in some models and explainer combinations. In the sufficiency metric, the synthetic reasoning not improves (by decreasing value) the explicit chain models, except for the RoBERTa model with LIME and IG. 

\section{Conclusion}


This work introduces a method for compliance detection using multi-hop inferences based on generated assurance cases using LLMs. We decompose long text requirements into claims, arguments, and evidence (CAE) of assurance cases. To assess the effectiveness of our approach, we first compared it against prior methods (RQ1). Formulating compliance detection as the NLI task yields competitive performance and better recall in most case. We then examined the relationship between model predictions and their explanations (RQ2). We observed that the model's correct prediction affects the explanation based on the Comprehensiveness metric. In contrast, the model's incorrect prediction relates more to the Sufficiency metric

In evaluating generalisation (RQ3), our explicit reasoning methods demonstrated superior performance on unseen requirements, highlighting the benefit of synthetic data from LLMs. Finally, we explored whether synthetic reasoning steps improve model faithfulness (RQ4). By comparing explicit vs implicit models, we found that incorporating intermediate reasoning chains increases model explainability based on the faithfulness metrics. Our results suggest that generating assurance cases with LLMs offers a promising direction for robust and explainable compliance detection.

\section{Limitations}  

Our study on compliance detection using Natural Language Inference (NLI) and multi-hop reasoning has several potential limitations that might be considered.


    Our evaluation relies on assurance cases generated by LLMs, which may introduce hallucinations, inconsistencies, or biases. While we propose evaluation metrics to measure structural coherence and instance-level consistency, these do not fully ensure logical correctness. 
    
    Our second experiment uses synthetic assurance case data due to the limited availability of expert-annotated real-world cases. This may affect the generalizability of our results when applied to actual regulatory audits and compliance verification

    While focusing on GDPR as a use case, our approach may not generalise seamlessly to other compliance domains such as HIPAA, ISO 27001, or SOX. Other regulatory texts may require domain-specific hyper-parameters, different architecture or transfer learning methods.

    The flat and structured metrics we propose measure structural coherence and textual consistency generated by LLMs. They do not aim to fully assess logical validity, factual correctness, or legal soundness. Future work can integrate our evaluation with formal verification or uncertainty quantification techniques to enhance reliability.

\section{Ethical Considerations}

Our second study experiments on synthetic assurance case data generated by large language models (LLMs) due to limited real-world datasets. While necessary for scalability, this introduces risks of hallucinated content, biased reasoning, or oversimplified logic that may not reflect actual regulatory processes.

The proposed system is not intended to replace legal or expert judgment, and we caution against its use in high-stakes compliance decisions without human oversight. Although we use GDPR as a case study, generalising to other regulatory domains (e.g., HIPAA or SOX) may require domain-specific adjustments.

We encourage responsible use of our methods in contexts where outputs are critically evaluated, and stress the importance of integrating expert validation and, where possible, formal verification.

\section{Acknowledgements}

This work has been funded by the European Commission under grant agreement No. 101120606, CERTIFAI. We thank Iker Lasa Ojanguren and Jose Arias Marin from FUNDACION TECNALIA RESEARCH \& INNOVATION for their valuable contributions to the discussion and technical knowledge for Generating Assurance Case with Large Language Models as part of the CERTIFAI project. We thank Sunanda Bose for the technical discussion on compliance detection. We would also like to thank Akriti Sharma for helping to improve the clarity of the paper's writing. This work has also benefited from the Experimental Infrastructure for Exploration of Exascale Computing (eX3), which is financially supported by the Research Council of Norway under contract 270053.

\appendix

\newpage

\section{Appendix}
\label{sec:appendix}



\subsection{NLI-Based Compliance Detection on GDPR-DPA Dataset experiments}

\paragraph{FT} All models are fine-tuned for six epochs, 32 batches, 1e-05 learning rate, and weight decay 0.01. The models are optimised with a 1e-05 learning rate, and the weights decay by 0.01.

\paragraph{LP-FT} All models are fine-tuned for three epochs in the probing phase and three in the fine-tuning phase. The batch size is 32. The models are optimised with a 1e-05 learning rate, and the weights decay by 0.01.

\paragraph{LoRA} All models are fine-tuned for six epochs, 32 batches, 1e-05 learning rate, and weight decay 0.01. The models are optimised with a 1e-05 learning rate, and the weights decay by 0.01. In LoRA parameters, we target these named parameters: ["attention.self.query", "attention.self.key", "attention.self.value", "attention.output.dense", "intermediate.dense"]

\paragraph{Zero-shot \& One-shot} For Decoder-based LMs, we used zero-shot and one-shot prompts to predict the entailment. The prompt template for zero-shot is:

\begin{tcolorbox}[colback=gray!10, colframe=black, title=Zero-shot Prompt, width=0.5\textwidth]
    "Below is a Natural Language Inference (NLI) task for compliance detection in general data privacy regulation domain."
    \newline
    "give answer in either 'entailment' or 'not entailment'"
    \newline
    f"Premise: \{premise\}"
    \newline
    f"Hypothesis: \{hypothesis\}"
    \newline
    "Answer:"
\end{tcolorbox}

The prompt template for one-shot is:

\begin{tcolorbox}[colback=gray!10, colframe=black, title=One-shot Prompt, width=0.5\textwidth]
    "Below is a Natural Language Inference (NLI) task for compliance detection in general data privacy regulation domain.
    \newline
    \newline"
    f"Example 1:\newline
    Premise: \{pos\_example[0]\}"
    \newline
    f"Hypothesis: \{pos\_example[1]\}"
    \newline
    "Answer: entailment"
    \newline
    \newline
    f"Example 2:
    \newline
    Premise: \{neg\_example[0]\}"
    \newline
    f"Hypothesis: \{neg\_example[1]\}"
    \newline
    "Answer: not entailment"
    \newline
    \newline
    f"Premise: \{premise\}"
    \newline
    f"Hypothesis: \{hypothesis\}"
    \newline
    "Answer:"
\end{tcolorbox}

\paragraph{Post-hoc Explanation Analysis}

\begin{figure}[ht]
    \includegraphics[width=0.5\textwidth]{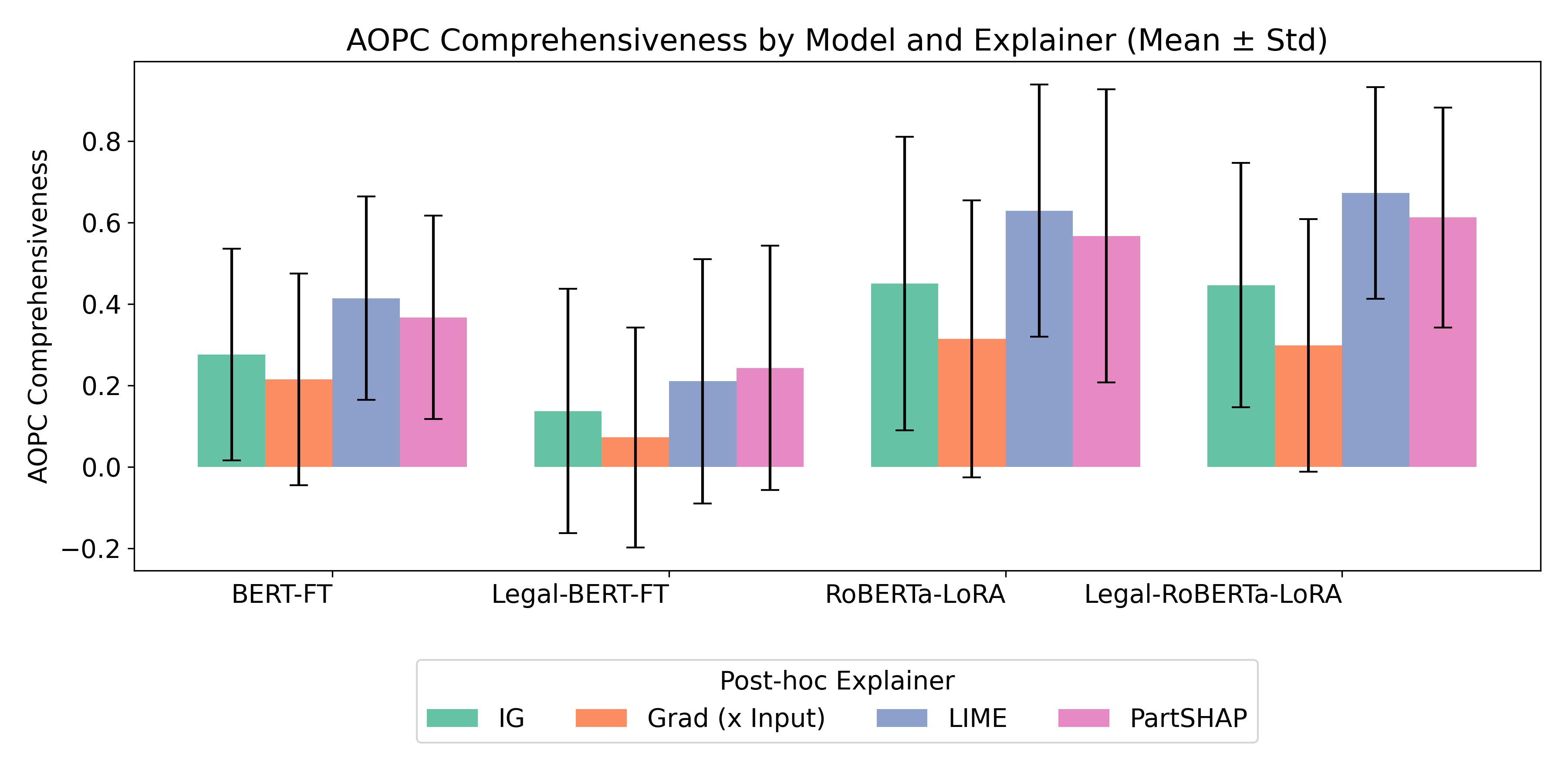}
    \vspace{1em} 
    \includegraphics[width=0.5\textwidth]{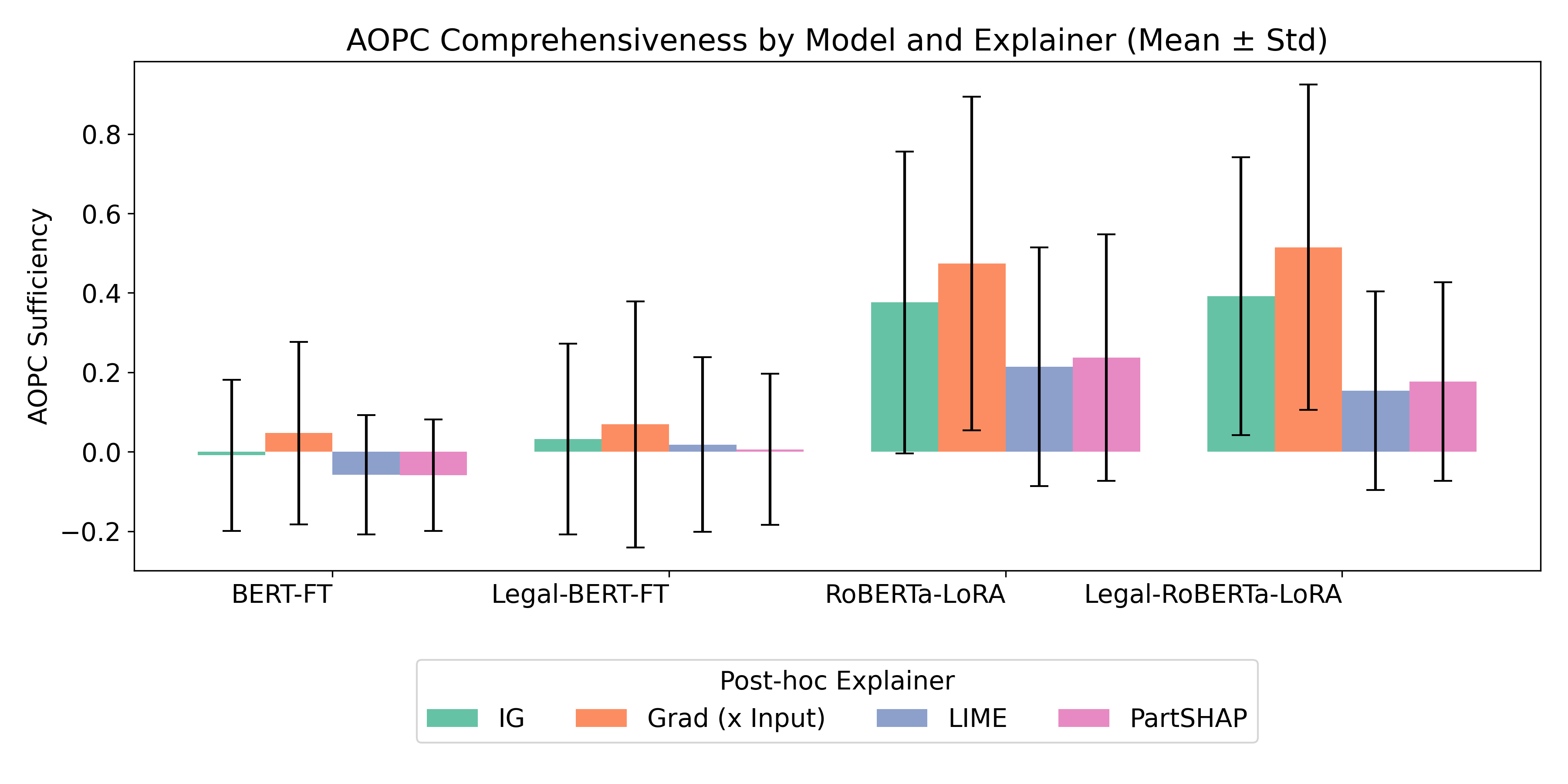}
    \caption{Comprehensiveness and Sufficiency Metrics by Model and Explainer}
    \label{fig:aopc_stacked}
\end{figure}



\begin{figure*}[ht]
    \centering
    \begin{subfigure}[b]{0.9\textwidth}
        \centering
        \includegraphics[width=0.9\textwidth]{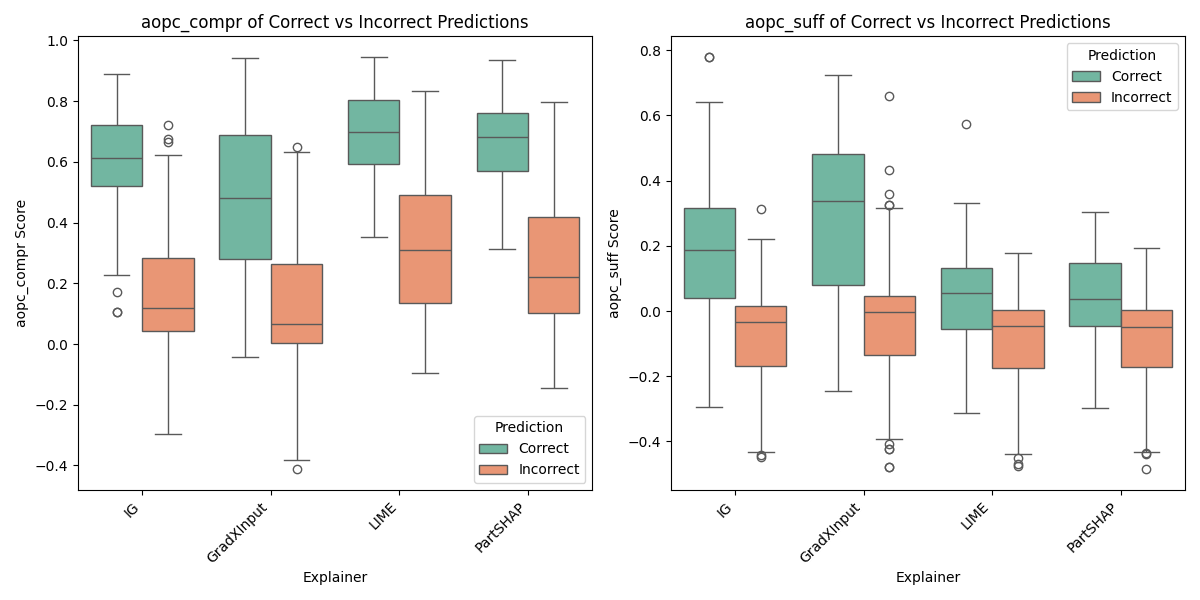}
        \caption{Score distribution of Correct vs Incorrect for BERT}
        \label{fig:aopc_cor_inc_bert}
    \end{subfigure}
    
    \vskip\baselineskip
    
    \begin{subfigure}[b]{0.9\textwidth}
        \centering
        \includegraphics[width=0.9\textwidth]{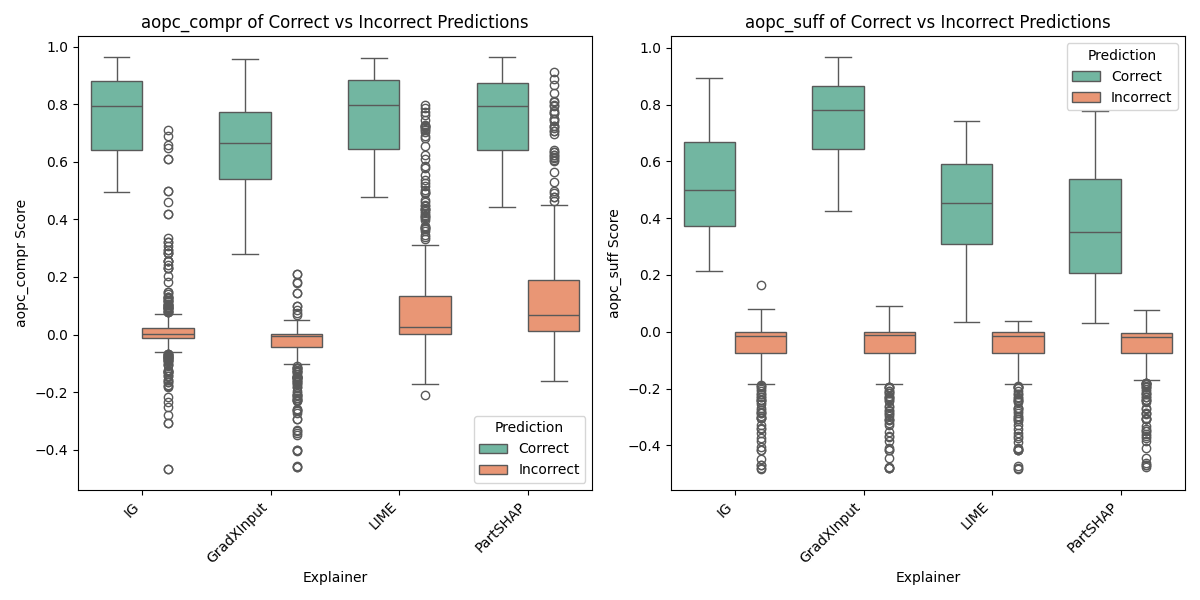}
        \caption{Score distribution of Correct vs Incorrect for Legal-BERT}
        \label{fig:aopc_cor_inc_legal_bert}
    \end{subfigure}
    \caption{Score distributions of Correct vs Incorrect for BERT and Legal-BERT}
    \label{fig:aopc_cor_inc_all-bert}
\end{figure*}



\begin{figure*}[ht]
    \centering
    \begin{subfigure}[b]{0.9\textwidth}
        \centering
        \includegraphics[width=0.9\textwidth]{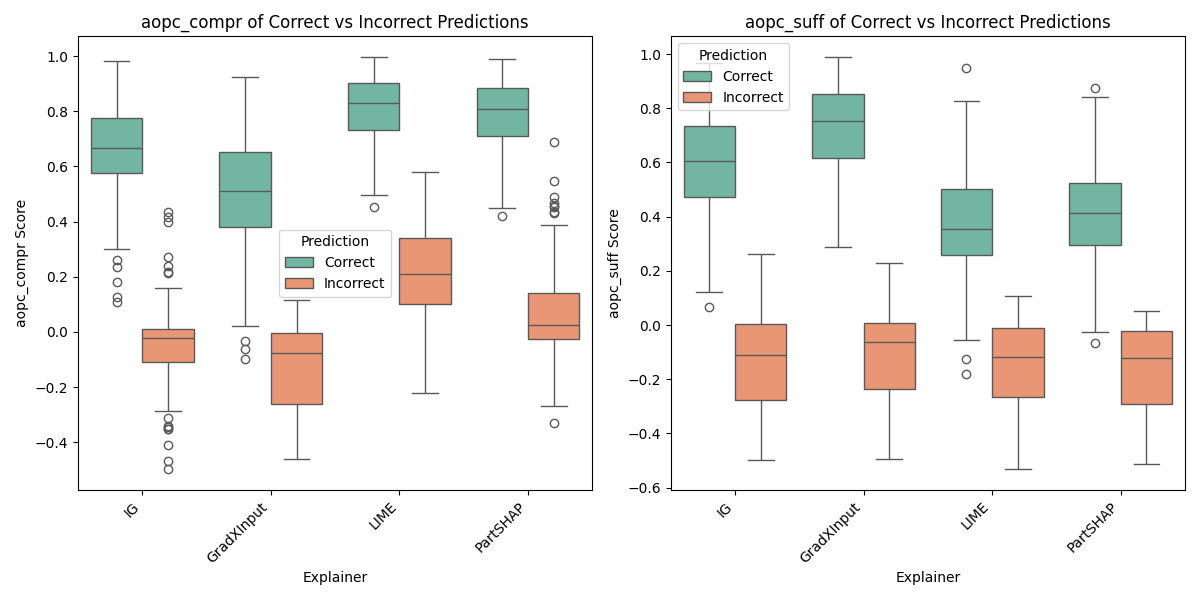}
        \caption{Score distribution of Correct vs Incorrect of RoBERTa-LoRA faithfulness metrics}
        \label{fig:aopc_cor_inc_roberta-lora}
    \end{subfigure}
    
    \vskip\baselineskip
    
    \begin{subfigure}[b]{0.9\textwidth}
        \centering
        \includegraphics[width=0.9\textwidth]{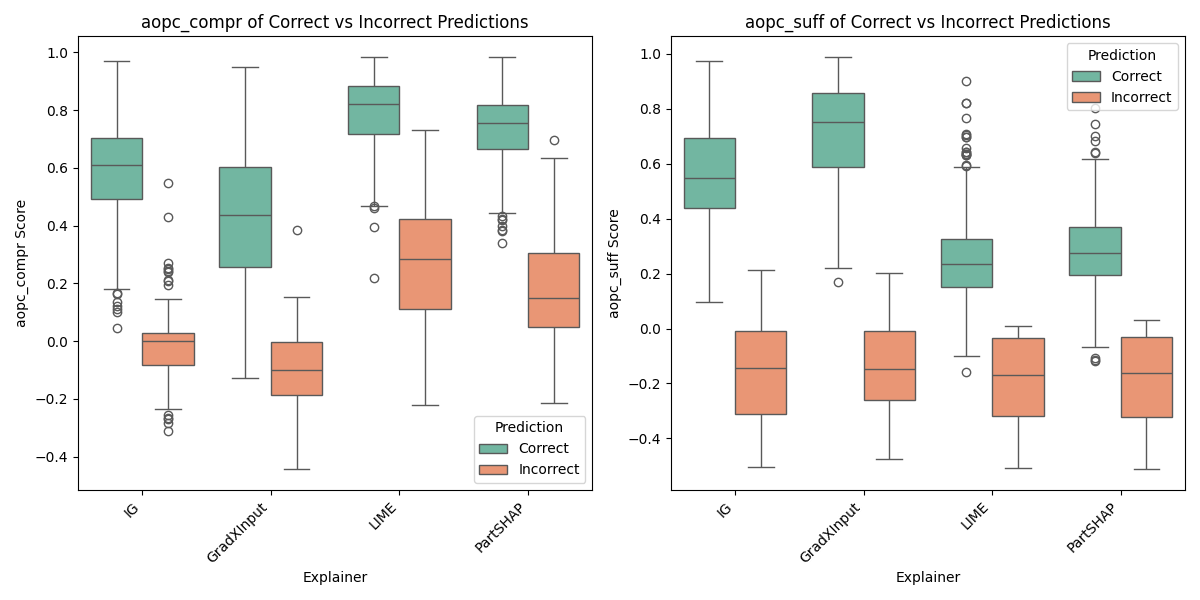}
        \caption{Score distribution of Correct vs Incorrect of Legal-RoBERTa-LoRA faithfulness metrics}
        \label{fig:aopc_cor_inc_legal-roberta-lora}
    \end{subfigure}
    \caption{Score distributions of Correct vs Incorrect for RoBERTa and Legal-RoBERTa}
    \label{fig:aopc_cor_inc_all-roberta}
\end{figure*}

\subsection{LLM Generated Data}

\paragraph{LLM Prompt}

In this paper, we use the following system and user role prompt:

\begin{tcolorbox}[colback=gray!10, colframe=black, title=LLM System Message, width=0.5\textwidth]
You are a legal expert in privacy and security issues. Your duty is to produce a thorough Data Processing Agreement Assurance Case from General Data Protection Regulation legal requirements. \{assurance\_case\_definition\}
\end{tcolorbox}



\begin{tcolorbox}[colback=gray!10, colframe=black, title=LLM Prompt, width=0.5\textwidth]

    \noindent The requirement is \textbf{\{requirement\_id\}}: \textbf{\{requirement\_name\}}

\vspace{1em} 

\noindent \textbf{Requirement Description}  

\{requirement\_description\}

\noindent \textbf{Rationale and Supplemental Guidance}  

\{requirement\_rationale\}

\vspace{1em} 

\noindent Give the output in \textbf{\{format\}}.
\end{tcolorbox}

\paragraph{Generated text to Assurance Case in JSON format Success rate}


\begin{table}[ht]
\caption{LLMs generation Error Summary}
\label{tab:error_summary}
\centering
\begin{tabular}{lr}
\toprule
\textbf{Metric}               & \textbf{Count} \\ \midrule
Number of error files         & 587            \\
Total files                   & 950            \\ \bottomrule
\end{tabular}
\end{table}

\begin{table}[ht]
\caption{Models with Error Output and Success Percentage (Total = 5 calls $\times$ 20 requirements = 100 cases)}
\label{tab:models_errors_success}
\centering
\begin{tabular}{lrr}
\toprule
\textbf{Model}               & \textbf{Err Cnt} & \textbf{Success \%} \\ \midrule
Qwen2.5-72B-Instruct         & 84                   & 16\%                        \\
Llama-3.1-8B-Instruct        & 92                   & 8\%                         \\
gemma-2-27b-it              & 95                   & 5\%                         \\
Llama-3.1-70B-Instruct       & 91                   & 9\%                         \\
phi-4                      & 5                    & 95\%                        \\
Qwen2.5-14B-Instruc t         & 32                   & 68\%                        \\
Llama-3.2-3B-Instruct        & 8                    & 92\%                        \\
Qwen2.5-7B-Instruct          & 0                    & 100\%                       \\
gemma-7b-it                & 95                   & 5\%                         \\
gemma-2-9b-it              & 85                   & 15\%                        \\ 
gpt35-turbo-4k             & 4                    & 4\%                         \\
gpt4o-8k                   & 6                    & 6\%                         \\
gpt4o                      & 2                    & 2\%                         \\
\bottomrule
\end{tabular}
\end{table}




We repeat five calls each in the assurance case generation process by 13 LLMs with 20 requirements. In total, there are 1,300 instances of JSON files. However, only around 25\% of files are valid JSON files in the first attempt. We make another call to the ChatGPT4o for each malformed JSON, which has a 100\% success rate in formatting JSON. Finally, we have 1000 valid Assurance Case JSON files, where 727 files are from Open Source LLMs and 273 from proprietary LLMs (gpt35-turbo-4k, gpt4o-8k, and gpt4o). Some failed outputs include copying the assurance case template and malformed structures, such as only containing MainClaim, missing SubClaim or containing no evidence element.

\subsection{Multi-hop Inference}

\paragraph{FT} All Multi-hop inference models are fine-tuned for 10000 steps, 32 batches, 1e-05 learning rate, and weight decays 0.01. The models are optimised with a 1e-05 learning rate, and the weights decay by 0.01.




\end{document}